\documentclass[10pt,twocolumn,letterpaper]{article}
\usepackage[pagenumbers]{cvpr}
\usepackage{graphicx}
\usepackage{amsmath}
\usepackage{amssymb}
\usepackage{booktabs}
\usepackage{bbding}
\usepackage{multirow}
\usepackage{float}
\usepackage{csquotes}
\usepackage{utfsym}
\definecolor{cvprblue}{rgb}{0.21,0.49,0.74}
\usepackage[pagebackref,breaklinks,colorlinks,allcolors=cvprblue]{hyperref}
\begin{document}
\title{Advancing Fine-Grained Visual Understanding with Multi-Scale Alignment \\in Multi-Modal Models}
\author{
    \textbf{Wei Wang\textsuperscript{* \rm 1,2}},
    ~\textbf{Zhaowei Li\textsuperscript{* \rm 2,3}},
    ~\textbf{Qi Xu\textsuperscript{\rm 2}},
    ~\textbf{Linfeng Li\textsuperscript{\rm 2}},
    ~\textbf{Yiqing Cai\textsuperscript{\rm 2}},
    ~\textbf{Botian Jiang\textsuperscript{\rm 2,3}},\\
    ~\textbf{Hang Song\textsuperscript{\rm 2}},
    ~\textbf{Xingcan Hu\textsuperscript{\rm 1}},
    ~\textbf{Pengyu Wang\textsuperscript{\rm 3}},
    ~\textbf{Li Xiao\textsuperscript{\dag \rm 1}}\\
    \textsuperscript{\rm 1}University of Science and Technology of China, \textsuperscript{\rm 2}ByteDance Inc, \textsuperscript{\rm 3}Fudan University \\
    {\tt wangweiii@mail.ustc.edu.cn, lizhaowei126@gmail.com} \\
}
\maketitle

\renewcommand{\thefootnote}{}
\footnotetext{*Equal contribution. Order is random.}
\footnotetext{\dag Corresponding author.}

%%%%%%%%% ABSTRACT
\raggedbottom
\begin{abstract}
Multi-modal large language models (MLLMs) have achieved remarkable success in fine-grained visual understanding across a range of tasks. However, they often encounter significant challenges due to inadequate alignment for fine-grained knowledge, which restricts their ability to accurately capture local details and attain a comprehensive global perception. While recent advancements have focused on aligning object expressions with grounding information, they typically lack explicit integration of object images, which contain affluent information beyond mere texts or coordinates. To bridge this gap, we introduce a novel fine-grained visual knowledge alignment method that effectively aligns and integrates multi-scale knowledge of objects, including texts, coordinates, and images. This innovative method is underpinned by our multi-scale fine-grained enhancement data synthesis pipeline, which provides over 300K essential training data to enhance alignment and improve overall performance. Furthermore, we present TinyGroundingGPT, a series of compact models optimized for high-level alignments. With a scale of approximately 3B parameters, TinyGroundingGPT achieves outstanding results in grounding tasks while delivering performance comparable to larger MLLMs in complex visual scenarios. The data and code will be released in \url{https://github.com/wwangweii/TinyGroundingGPT.git}.

\end{abstract}

\section{Introduction}
\label{sec:intro}

Recent advancements in multi-modal large language models (MLLMs) have showcased remarkable capabilities in multi-modal understanding, reasoning, and interaction, garnering unprecedented attention~\cite{touvron2023llama,chen2023shikra,peng2023kosmos,bai2023qwen,wang2023cogvlm,chen2024far}. MLLM research in fine-grained visual understanding has advanced significantly, particularly through early contributions from Shikra~\cite{chen2023shikra} and Kosmos-2~\cite{peng2023kosmos} in textually formatting positional vocabularies or object coordinates. Subsequent studies aimed at improving model performance primarily focused on common strategies, including parameter enlargement~\cite{chen2023shikra, peng2023kosmos, li2024groundinggpt, bai2023qwen} and dataset enrichment~\cite{chen2023sharegpt4v, bai2023qwen, wang2023cogvlm, chen2024far}. Additionally, there is a growing interest in developing efficient, smaller fine-grained MLLMs~\cite{li2024mini, hu2024minicpm, yao2024minicpm, zhu2023minigpt,zhou2024tinyllava} for real-world applications. Regardless of the methods used, the core of fine-grained models lies in achieving better alignment between object texts and visual features, encompassing both coordinate and semantic information.

\begin{figure}[t]\small
    \centering
    \includegraphics[width=1.0\columnwidth]{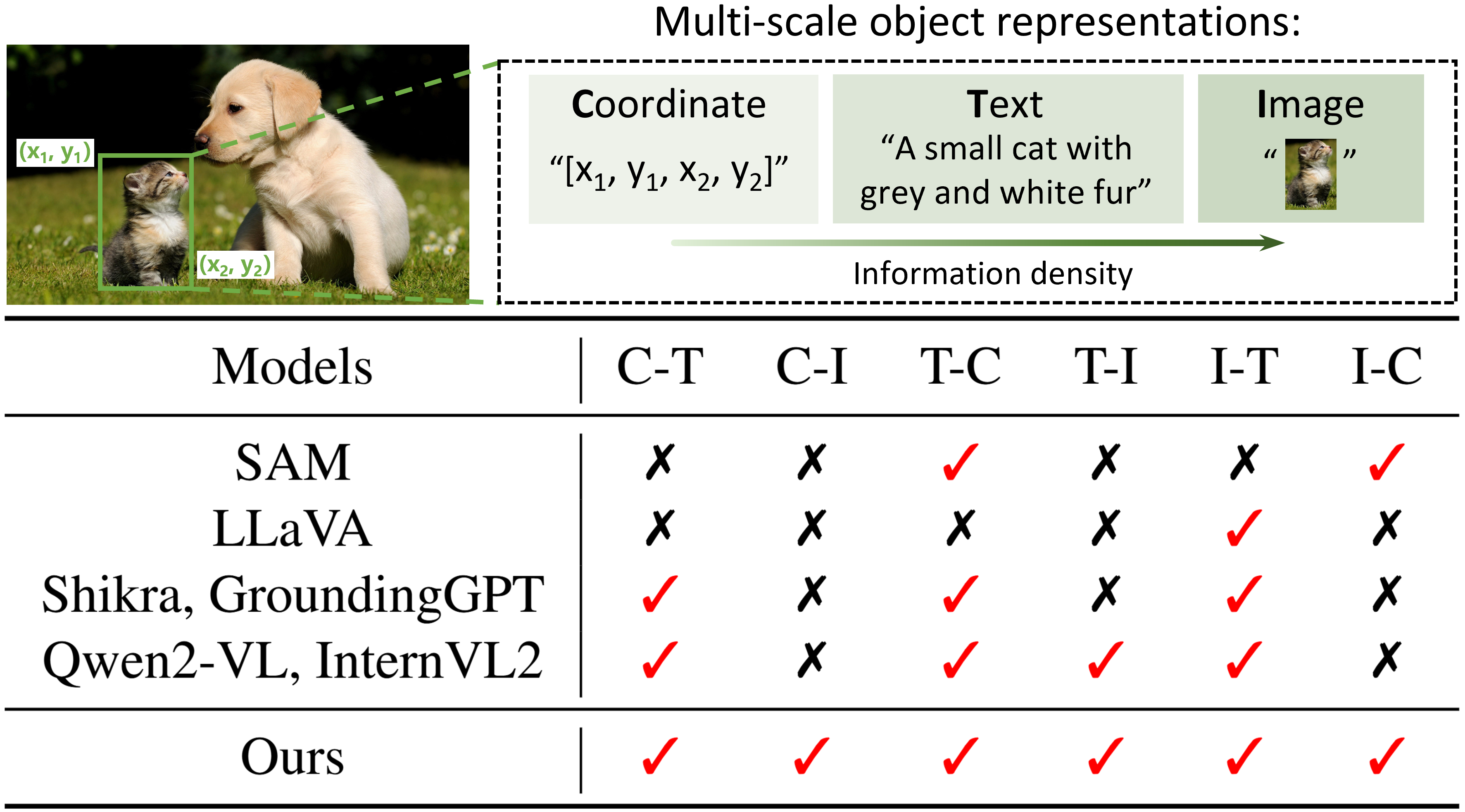}
    \caption{The comparison of alignment for multi-scale object representations. The C, T, I denote object coordinates, texts and images respectively. The \enquote{X-Y} denote MLLMs handle input \enquote{X} and output \enquote{Y}.}
    \label{fig:comaprison}
\end{figure}

\begin{figure*}[!ht]
\centering
\includegraphics[width=2\columnwidth]{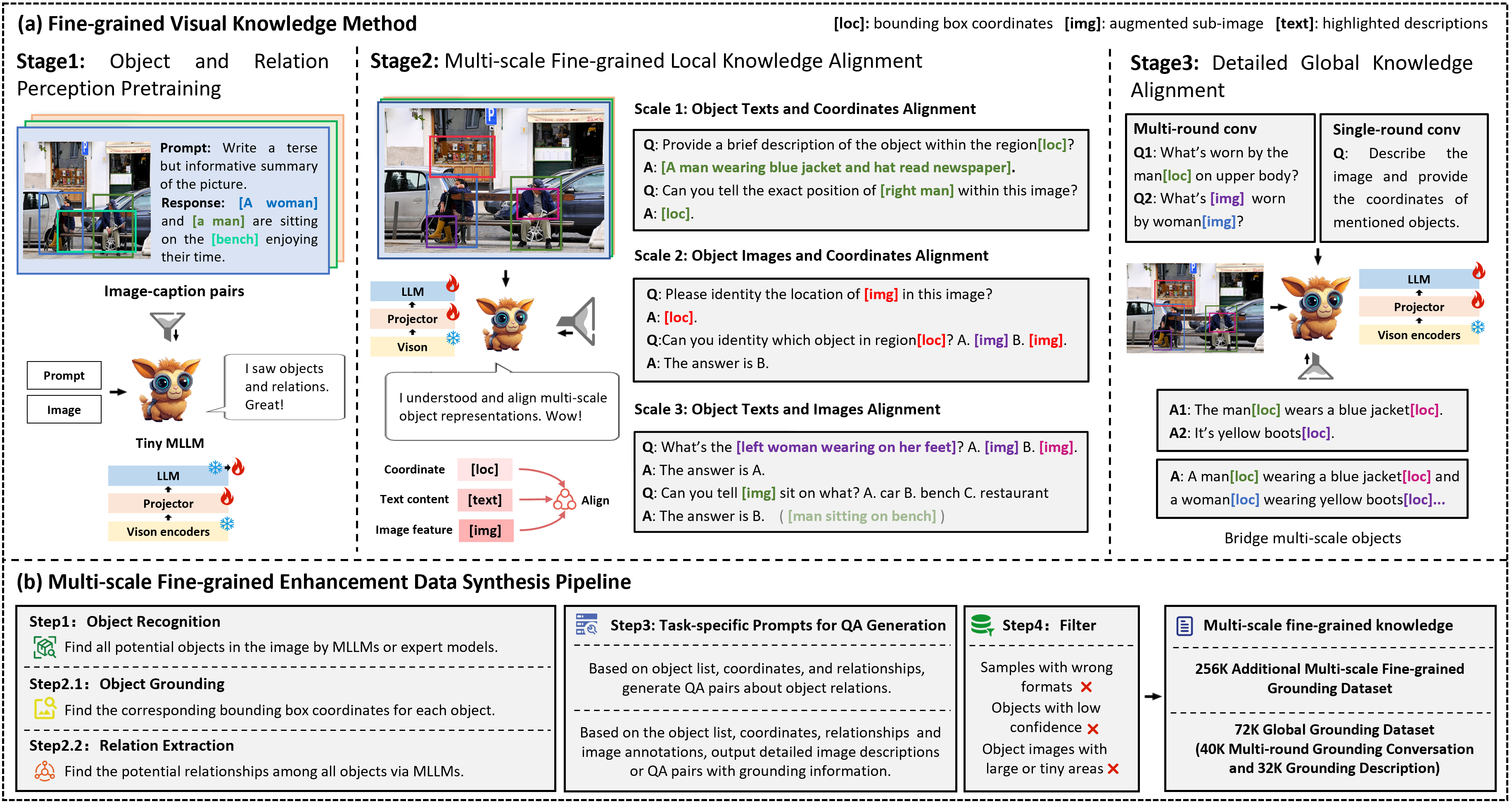}
\caption{Illustration of the proposed multi-modal fine-grained visual knowledge alignment method. It adopts a three-stage training strategy that progresses from easy to hard and the multi-scale fine-grained enhancement data synthesis pipeline constructs over 300K fine-grained alignment data.}
\label{fig1}
\end{figure*}

While effective, these methods face a significant challenge, i.e., the lack of fine-grained alignments. Visual objects typically encompass multi-scale representations with varying levels of information, including coordinates, texts, and images, as illustrated in Fig.~\ref{fig:comaprison}. In this context, coordinates provide low-level object grounding information, texts offer primary descriptions that may not capture every detail, and images convey high-level information that extends beyond words. Most fine-grained models~\cite{chen2023shikra, you2023ferret, li2024groundinggpt} primarily focus on alignments between object texts and coordinates (i.e., T-C and C-T), often neglecting direct interactions with object images. Although recent models like Qwen2-VL~\cite{bai2023qwen} and InternVL2~\cite{chen2024far} can process multiple image inputs and understand relationships between the main image and object images (T-I), they still struggle to establish explicit alignments among object texts, coordinates, and images. This limitation can lead to hallucinations and insufficient grounding capabilities~\cite{chen2024unified}.

To achieve high-level alignments and integrate multi-granularity knowledge, as illustrated in Fig.~\ref{fig1}(a), we introduce a fine-grained visual knowledge alignment method that effectively aligns object texts, coordinates, and images across multiple scales. Our method adopts a three-stage training strategy that progresses from easy to hard: 1) Object and Relation Perception Pretraining: To develop a foundational understanding of object texts and images, we implement a progressive training approach for MLLMs based on a pretrained LLM. 2) Multi-scale Fine-grained Local Knowledge Alignment: To attain fine-grained visual understanding and share multi-scale object knowledge, we conduct data-driven high-level alignments among object text descriptions, bounding box coordinates, and image features. 3) Detailed Global Knowledge Alignment: To enhance the model's global understanding by integrating fine-grained knowledge, we guide the MLLMs to bridge different objects with multi-scale representations. To support this method, we propose a multi-scale fine-grained enhancement data synthesis pipeline (see Fig.~\ref{fig1}(b)) that constructs alignment data from both local and global perspectives. Leveraging this framework, we propose TinyGroundingGPT, which requires less storage for deployment while outperforming larger models across multiple benchmarks, particularly in grounding tasks. Our contributions can be summarized as follows:
\begin{itemize}
\setlength{\itemsep}{0pt}
\setlength{\parsep}{0pt}
\setlength{\parskip}{0pt}
\item We introduce a fine-grained visual knowledge alignment method that enables the model to progressively enhance its fine-grained visual understanding through both global and local multi-scale object alignments.
\item We develop a multi-scale fine-grained enhancement data synthesis pipeline that leverages open-source datasets and advanced models to generate over 300K essential training data for fine-grained alignment.
\item We present TinyGroundingGPT, a series of compact models (1.5B and 3B parameters) that excel in multi-modal understanding and grounding capabilities, achieving performance comparable to that of larger 7B MLLMs.

\end{itemize}

\section{Related works}
\label{sec:relatedwork}

\paragraph{Multi-modal Large Language Models}
Recent advancements in large language models (LLMs) like ChatGPT and LLaMA~\cite{touvron2023llama} have significantly propelled the development of multi-modal large language models. Notable proprietary models, such as GPT-4V~\cite{OpenAI2023}, have showcased the potential of multi-modal capabilities in visual tasks. Early open-source initiatives include BLIP-2~\cite{li2023blip}, MiniGPT-4~\cite{zhu2023minigpt}, and LLaVA~\cite{liu2024visual}, which leverage pre-trained LLMs and excel in tasks like visual question answering. Subsequent efforts, including Qwen-VL~\cite{bai2023qwen}, InternVL~\cite{chen2024far}, and MiniCPM-V~\cite{yao2024minicpm}, have further enhanced model capabilities by introducing dynamic resolution, expanding training data, and incorporating reinforcement learning, achieving impressive results in optical character recognition (OCR) and grounding, while also enhancing the credibility of model responses.

Despite these advancements, the large number of parameters in MLLMs incurs extremely high costs in training and deployment, limiting their widespread application. Many studies have explored how to build more lightweight LLMs,
such as Mini-Gemini~\cite{li2024mini}, MobileVLM~\cite{chu2024mobilevlm}, MiniCPM-V~\cite{yao2024minicpm}, etc. They are based on lightweight LLMs, combined with optimized structures or training strategies, enabling to have performance comparable to that of larger models. 
~\cite{hsieh2023distilling,wang2024qcrd,shu2024llava} have explored how to distill capabilities from larger models, so that small models can obtain complex reasoning abilities. 

\paragraph{Fine-grained Multi-modal Models}\setlength{\parskip}{0em}
Recent research has increasingly focused on multi-modal language models capable of fine-grained understanding, which can be applied to complex tasks such as grounding and OCR. Methods like Shikra~\cite{chen2023shikra} and Kosmos-2~\cite{peng2023kosmos} enhanced the visual grounding capabilities of multi-modal large language models by constructing datasets that include coordinate information, often by transforming visual task datasets into an instruction-following format using templates. Other approaches integrated additional visual components, such as GLaMM~\cite{rasheed2024glamm} and LLaVA-Grounding~\cite{zhang2023llava}, or extracted regional features as supplementary inputs, as seen in Ferret~\cite{you2023ferret}, NExT-Chat~\cite{zhang2023next}, and GPT4RoI~\cite{zhang2023gpt4roi}. GroundingGPT~\cite{li2024groundinggpt} extended to support grounding tasks across multiple modalities. Moreover, some initiatives sought to broaden the capabilities of MLLMs for various visual tasks, such as VisionLLMv2~\cite{wu2024visionllm} and UnifiedMLLM~\cite{li2024unifiedmllm}, which facilitate tasks like image editing and image segmentation. To enhance performance on fine-grained tasks, models like LLaVA-UHD~\cite{xu2024llava} and InternVL~\cite{chen2024far} have explored dynamic high-resolution techniques, improving results in areas like OCR. However, these models often lack systematic alignments among object texts, coordinates, and images, limiting the full integration of these multi-scale representations.

\begin{figure*}[thp]
\centering
\includegraphics[width=1.6\columnwidth]{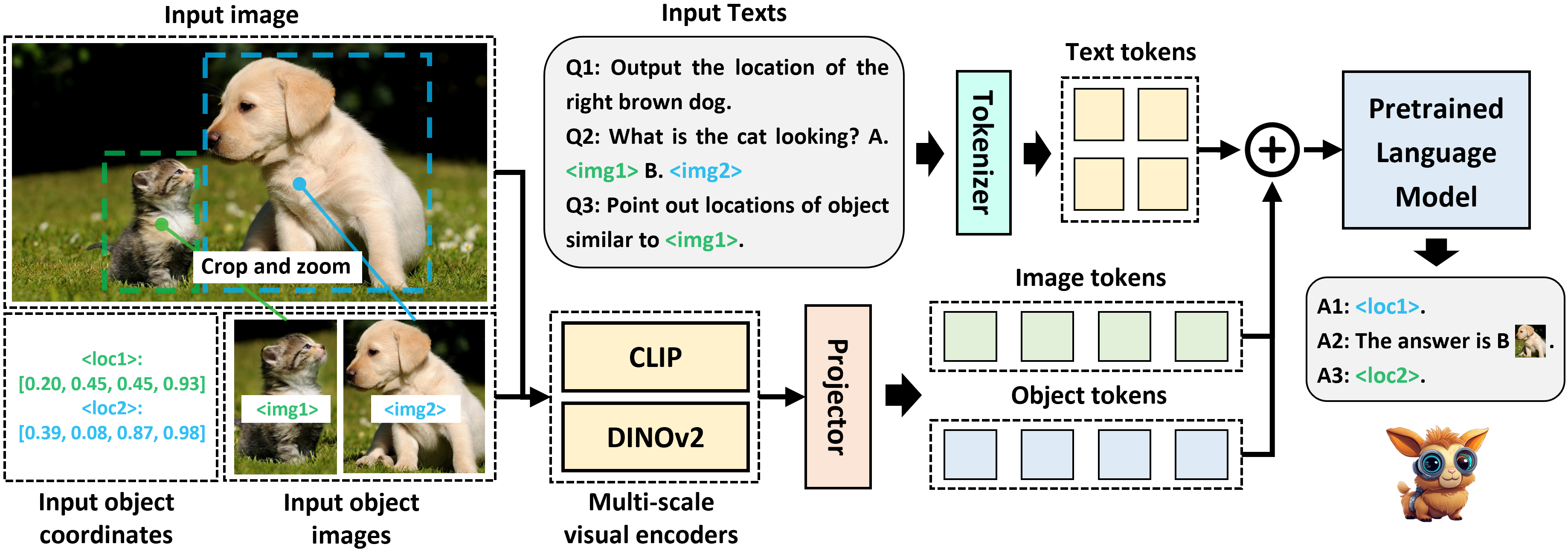}\\
\caption{The model architecture of our proposed TinyGroundingGPT. It utilizes multi-scale visual encoders and supports queries regarding different object representations. Object images are cropped and zoomed from the input image according to the input coordinates.}
\label{model}
\end{figure*}
\section{Method}
\label{sec:method}

In this paper, we first introduce a novel fine-grained visual knowledge alignment method that harnesses the potential of MLLMs by aligning object texts, coordinates, and images across multiple scales, as shown in Fig.~\ref{fig1}(a). Our method consists of three training stages that progress from easy to hard: (a) Object and Relation Perception Pretraining, which enables the model to understand multimodal inputs, identifying objects in images and their interrelations. (b) Multi-scale Fine-grained Local Knowledge Alignment by which the model is guided to achieve multi-scale, fine-grained alignments, accommodating diverse inputs such as object texts, coordinates, and images. (c) Detailed Global Knowledge Alignment which focuses on model training for global alignment and understanding, further integrating fine-grained information and bridging different objects with multi-scale representations. To support this high-level alignment, we then propose a multi-scale fine-grained enhancement data synthesis pipeline, as illustrated in Fig~\ref{fig1}(b), which generates multi-scale alignment datasets from both global and local perspectives. Building on this framework, we propose TinyGroundingGPT, which requires less storage for deployment while outperforming larger parameter models across multiple benchmarks, particularly in hallucination evaluation and grounding tasks.

\subsection{Fine-grained Visual Knwoledge Alignment}
We elaborate the three training stages in our fine-grained visual knowledge alignment method below.\\

\noindent\textbf{Object and Relation Perception Pretraining } In this stage, we aim for the model to comprehend multi-modal inputs, recognizing the objects present in the image and the relationships among them, which forms the foundation for subsequent reasoning and grounding tasks. Throughout the training process, we initially keep the LLM and encoder frozen, training only the projector to connect the text and image semantic spaces. Subsequently, we train both the LLM and the projector to enhance the understanding of objects and their relationships. We utilize LLaVA-Pretrain-595k~\cite{liu2024visual} and each sample is
accompanied by a sampled instruction that requires
the model to provide a concise description of the image.\\

\noindent\textbf{Multi-scale Fine-grained Local Knowledge Alignment } After the initial training stage, where the model learns to recognize objects and their relationships, it still lacks the grounding capability to accurately locate these objects in images and to integrate different representations of a single object. In this stage, we therefore train the model to achieve fine-grained alignments among object texts, coordinates, and images, fully sharing their multi-scale knowledge for each representation. We utilize original visual grounding datasets such as RefCOCO~\cite{kazemzadeh2014referitgame}, RefCOCO+~\cite{kazemzadeh2014referitgame}, RefCOCOg~\cite{mao2016generation} and Visual Genomes~\cite{krishna2017visual}, along with a developed multi-scale fine-grained enhancement data synthesis pipeline (details provided in the following subsection) to construct a fine-grained grounding dataset. The instances in the training data can be categorized into three classes:
\begin{itemize}
\setlength{\itemsep}{0pt}
\setlength{\parsep}{0pt}
\setlength{\parskip}{0pt}
  \item Object \textbf{Texts} and \textbf{Coordinates} Alignment: The model refers to corresponding coordinates for a given object text description or describes a region based on input coordinates. 
  \item Object \textbf{Images} and \textbf{Coordinates} Alignment: Given an augmented object image, the model identifies its location within the image. When provided with coordinates, the model selects the most relevant object images.
  \item Object \textbf{Texts} and \textbf{Images} Alignment: The model selects the most relevant augmented object image based on the input question or answers inquiries about the relationships involving augmented object images.
\end{itemize}

Throughout the training process, we train both the LLM and the projector. Afterwards, the model can effectively perform fine-grained image understanding by achieving high-level alignments among object texts, images, and coordinates, while sharing multi-scale knowledge across each representation. Additionally, the forms of input objects to the model have been expanded.\\

\noindent\textbf{Detailed Global Knowledge Alignment }Despite achieving a fine-grained understanding of multi-modal data in the previous stage, the model lacks systematic training for global image comprehension and the ability to connect different objects with varied representations. Specifically, in the previous stage, only the representations of individual objects in each training sample were aligned. In this stage, our goal is to further align and integrate multiple objects within a single image input to enhance global knowledge learning. To achieve this, in addition to utilizing common image annotation datasets for instruction tuning, including LLaVA-v1.5-mix665k~\cite{liu2024visual} and ShareGPT4V~\cite{chen2023sharegpt4v}, we construct a global grounding dataset with high-level fine-grained alignments based on Flickr30K Entities~\cite{plummer2015flickr30k}: 1) Multi-round Grounding Conversation Data: This dataset guides the model to achieve a global understanding of the image through multi-round conversations, requiring it to combine fine-grained knowledge and thoroughly explore the relationships among different representations of various objects. 2) Grounding Description Data: This dataset prompts the model to provide a detailed description of the image to connect multi objects in one-round conversations, where the generated object texts are enhanced with coordinates to confirm their existence and effectively integrate grounding information.

This method enables us to leverage the fine-grained grounding alignment learned in the second stage to enhance the model's global grounding alignment. Additionally, we train both the LLM and the projector in this stage.

\subsection{Multi-scale Fine-grained Enhancement Data
Synthesis Pipeline}\label{datapipeline}
As shown in Fig.~\ref{fig1}(b), we develop a multi-scale fine-grained enhancement data synthesis pipeline, and construct a multi-scale fine-grained grounding dataset (in Stage2) as well as a global grounding dataset (in Stage3). Specifically, given an image, we perform the following steps:\\
\noindent\textbf{Object Recognition} We employ expert models or MLLMs for object detection in the input images, generating a list of identified objects, referred to as $L_1$. It is crucial to ensure that all listed objects exist in the image to avoid introducing hallucination in the subsequent data generation process. A prompt example for GPT-4V is provided in Appendix Fig.\ref{prompt1}.\\
\noindent\textbf{Object Grounding} In addition to the object text and coordinate pairs in the original datasets such as RefCOCO, we utilize the object list $L_{1}$ and the corresponding image to apply grounding models for each object in order to obtain bounding box coordinates. In this paper, we employ GroundingDINO~\cite{liu2023grounding} to locate the objects in the list and filter out those with low confidence, resulting in an object bounding box dictionary $S_{1}$. \\
\noindent\textbf{Relationship Extraction} To explore the relationships among objects and provide more material for subsequent QA generation, we prompt GPT-4V to extract potential relationships among the objects. As seen in Appendix Fig.\ref{prompt1}, given the object list $L_{1}$ and image, we obtain a list $L_{2}$ that consisits of triple in the format $(<object1>, <relation1>, <object2>)$.\\
\noindent\textbf{QA Generation} Based on above $L_{1}$, $S_{1}$ and $L_{2}$, we use task-specific prompts for GPT to generate different kinds of datasets (we provide case examples in Appendix Figs.~\ref{QApairs} and ~\ref{QApairs2}): (1) 256K Additional Multi-scale Fine-grained Grounding Dataset: compared to previous works~\cite{li2024groundinggpt} that focused solely on the alignment between object texts and coordinates, we enhance the alignment format by constructing an additional multi-scale, fine-grained dataset of instances based on objects and their relationships. This dataset incorporates object images, texts, and coordinates, facilitating more fine-grained image understanding and multi-scale alignment. Specifically, in addition to QAs that involve describing objects given coordinates or locating objects based on descriptions, we prompt GPT to generate QAs about relationships and mark the objects in the questions. We then replace these objects with augmented object images in the questions or options. Details can be seen in Appendix Fig.~\ref{prompt2}. (2) 72K Global Grounding Dataset: To enhance the global alignment and bridge objects with various representations, we construct two kinds of datasets: 1) 40K Multi-round Grounding Conversation: This dataset includes multi-turn dialogue formats, focusing on point-to-point questions about local details. 2) 32K Grounding Description: This dataset features single-turn dialogue formats, emphasizing an understanding of overall image descriptions with fine-grained grounding information. We provide prompts in Appendix Fig.~\ref{prompt3} and Appendix Fig.~\ref{prompt4}.\\
\noindent\textbf{Filter} We filter out QAs that contain object images with areas that are either too large or too small, as well as those with high Intersection over Union (IoU) among object images in the options. Additionally, we exclude QAs related to objects with low confidence or those with an excessive number of bounding boxes. This is to avoid low-resolution noise or image reference ambiguity.

\subsection{TinyGroundingGPT}
Using our proposed alignment method and synthesis data, we train TinyGroundingGPT to demonstrate the effectiveness of our proposed method.
Fig~\ref{model} illustrates the overall architecture of the TinyGroundingGPT. Images in various formats are processed through multi-scale vision encoders to extract features. Specifically, we extract image features using the pre-trained visual encoder ViT-L/14~\cite{radford2021learning} and pre-trained Dinov2-L/14~\cite{oquab2023dinov2}, concatenating them to obtain image features that incorporate both the global perception capabilities of CLIP and the local fine-grained understanding of DINOv2~\cite{jiang2023clip}. These features are then mapped to the LLM embedding space using an MLP. Note that in our proposed TinyGroundingGPT, the input supports both global image and object images, each represented by different special tokens: $<image>$ and $<object>$. These object images are cropped and zoomed from the global image according to the corresponding coordinates. We also support the input and output of object bounding box coordinates $<loc>$, represented in the text format $[x1, y1, x2, y2]$, with values ranging from 0.000 to 1.000.

\begin{table*}[ht]\small
    \centering
    \begin{tabular}{ccc|ccc|ccc|cc|c}
    \toprule
     \multirow{2}{*}{Type}  & \multirow{2}{*}{Model} & \multirow{2}{*}{LLM Size} & \multicolumn{3}{c}{RefCOCO} & \multicolumn{3}{c}{RefCOCO+} & \multicolumn{2}{c}{RefCOCOg} &
     \multirow{2}{*}{Avg}\\
    \cline{4-11}  & & & val & testA & testB & val & testA & testB & val & test & \\
    \midrule
    \multirow{3}{*}{Specialist} & UNITER     & -  & 81.41 & 87.04 & 74.17 & 75.90 & 81.45 & 66.70 & 74.02 & 68.67 & 76.17\\
     & MDETR      & -  & 86.75 & 89.58 & 81.41 & 79.52 & 84.09 & 70.62 & 81.64 & 80.89 & 81.81\\
     & UniTAB     & -  & 86.32 & 88.84 & 80.61 & 78.70 & 83.22 & 69.48 & 79.96 & 79.97 & 80.89\\
    \midrule
    \multirow{7}{*}{Generalist} & KOSMOS-2   & 1.6B  & 52.32 & 57.42 & 47.26 & 45.48 & 50.73 & 42.24 & 60.57 & 61.65 & 52.21\\ 
     & Shikra     & 7B  & 87.01 & 90.61 & 80.24 & 81.60 & \underline{87.36} & 72.12 & 82.27 & 82.19 & 82.93\\
     & NExT-Chat* & 7B & 85.50  & 90.00  & 77.90  & 77.20  & 84.50  & 68.00  & 80.10  & 79.80 & 80.38\\
     & Ferret*    & 7B & 87.49 & 91.35 & 82.45 & 80.78 & \textbf{87.38} & 73.14 & \textbf{83.93} & \textbf{84.76} & \underline{83.91}\\

     & GroundingGPT & 7B & \underline{88.02} & \underline{91.55} & \underline{82.47} & \underline{81.61} & 87.18 & \underline{73.18} & 81.67 & 81.99 & 83.46 \\  
     & $\text{InternVL2}^{+}$ & 2B & 82.3 & 88.2 & 75.9 & 73.5 & 82.8 & 63.3 &  77.6 & 78.3 & 77.74\\
     & $\text{Qwen2-VL}^{+}$ & 2B & 87.6 & 90.6 & 82.3 & 79.0 & 84.9 & 71.0 &  81.2 & 80.3 & 82.11\\ \midrule
    \multirow{2}{*}{Generalist} & \multirow{2}{*}{TinyGroundingGPT} & 3B & \textbf{89.16} & \textbf{92.24} & \textbf{85.38} & \textbf{81.70} & 87.16 & \textbf{75.09} & \underline{83.27} & \underline{84.08} & \textbf{84.76}\\

    & & 1.5B & 86.76 & 90.42 & 81.81 & 78.86 & 84.65 & 70.24 & 79.88 & 80.04 & 81.58\\
    \bottomrule
    \end{tabular}
    \caption{Performance comparison on the referring expression comprehension(REC) task. "*" indicates that the model employs additional image region perception modules and "+" indicates that the model uses dynamic high-resolution. The best results are highlighted in bold, while the second-best results are underlined.}
    \label{tab:refcoco}
\end{table*}

\section{Experiments}
\label{sec:experiments}
\subsection{Experimental Setup}
We employ Qwen2.5-3B and Qwen2.5-1.5B~\cite{qwen2.5} as the language models for our TinyGroundingGPT. During the training process, all images were padded to a square shape and resized to a resolution of 336 × 336. For more details on the hyper-parameter settings, training processes and datasets, please refer to the Appendix~\ref{Implementation} and ~\ref{Dataset}.

\subsection{Image Grounding Evaluation}
To evaluate the image grounding capability of TinyGroundingGPT, we conducted experiments on the Reference Expression Understanding (REC) task, which involves locating the bounding box for a given text reference. Our experiments utilized three datasets: RefCOCO, RefCOCO+, and RefCOCOg. We compared TinyGroundingGPT against various baseline models, including end-to-end multi-modal models like UNITER~\cite{chen2020uniter}, MDETR~\cite{kamath2021mdetr}, and UniTAB~\cite{yang2022unitab}, as well as LLM-based models such as KOSMOS-2, Shikra, NExTChat, Ferret, and GroundingGPT. Additionally, smaller models like InternVL2 and Qwen2-VL were included. We used a unified prompt formatted as \enquote{Output the coordinate of $\mathrm{<exp>}$}, where \enquote{$\mathrm{<exp>}$} represents the reference expression. As shown in Table~\ref{tab:refcoco}, TinyGroundingGPT demonstrates strong performance across all datasets, even with smaller LLM sizes (3B and 1.5B), matching or exceeding the performance of specialized fine-tuned models and larger MLLMs with additional image perception modules. Notably, the 3B model achieved state-of-the-art results on several benchmarks, attaining the highest average accuracy. Furthermore, TinyGroundingGPT-1.5B showed comparable grounding results, outperforming Next-Chat-7B on nearly all test sets.

\begin{table*}[th]\small
    \centering
    \renewcommand\arraystretch{0.95}
    \setlength{\tabcolsep}{1mm}  {
    \begin{tabular}{cc|ccccccc}
    \toprule
    Models & LLM Size & $\text{VQA}^{\text{v2}}$ & GQA & $\text{SQA}^{\text{I}}$ & POPE & $\text{MME}^{P}$ & MMB & $\text{LLaVA}^{\text{W}}$\\
    \midrule
    BLIP-2      & 13B    & 41.0  & 41     & 61    & 85.3  & 1293.8    & -  & 38.1   \\
    InstructBLIP  & 7B     & -     & 49.2  & 60.5  & -     & -         & 36 & 60.9   \\
    InstructBLIP & 13B    & -     & 49.5   & 63.1   & 78.9  & 1212.8    & -  & 58.2   \\ 
    Shikra     & 13B    & 77.4   & -    & -       & -     & -         & 58.8 & - \\ 
    LLaVA-1.5    & 7B     & 78.5  & 62.0   & 66.8   & 85.9  & \textbf{1510.7}   & 64.3 & 63.4  \\ 
    
    GroundingGPT  & 7B     & \underline{78.7}  & 62.1   & -   & 87.4  & 1454.2    & 63.8 & \textbf{70.9} \\ 
    Qwen-VL-Chat  & 7B     & 78.2  & -   & 68.2   & -  & \underline{1487.5}    & 60.6 & - \\ 
    \midrule
    $\text{MiniCPM-V-2}^{+}$   & 2.8B  & - & - & - & 87.8 & - & \textbf{69.6} & \underline{69.2}\\
    
    $\text{InternVL-2}^{+}$  & 2B  & - & 61.0 & - & \textbf{88.3} & 1439.6 & - & 62.5\\
    LLaVA-Phi & 2.7B  & 71.4 & - & 68.4 & 86.7 &  1335.1 & 59.8 & -\\ 
    TinyLLaVA & 2.7B  & 77.7 & 61.0 & \underline{70.1} & 86.3 &  1437.3 & \underline{68.3} & 67.1\\ 
    \midrule
    \multirow{2}{*}{TinyGroundingGPT} & 3B & \textbf{79.3} & \textbf{63.3}  & \textbf{70.3} & \underline{87.9} & 1423.2 & 66.4 & 67.5\\

     & 1.5B & 77.9  & \underline{62.2} & 63.1 & 87.6 & 1392.4 & 64.2 & 65.3\\
    \bottomrule
    \end{tabular}}
    \caption{Comparison of MLLMs on image understanding benchmarks. Benchmark names are abbreviated due to space limits. VQA-v2~\cite{goyal2017making}; GQA~\cite{hudson2019gqa}; $\text{SQA}^{\text{I}}$:ScienceQA-IMG~\cite{lu2022learn}; POPE~\cite{li2023evaluating}; MME~\cite{fu2023mme}; MMB:MMBench~\cite{MMBench}; $\text{LLaVA}^{\text{W}}$: LLaVA-Bench (In-the-Wild)~\cite{liu2024visual}. "+" indicates that the model uses dynamic high-resolution.}
    \label{tab:image_understanding}
\end{table*}

\begin{table*}[th]\small
    \centering
    \renewcommand\arraystretch{0.95}
    \setlength{\tabcolsep}{0.4mm}{
    \begin{tabular}{cc|ccc|ccc|ccc}
    \toprule
    \multirow{2}{*}{Models} & \multirow{2}{*}{LLM Size} & \multicolumn{3}{c}{Random} & \multicolumn{3}{c}{Popular} & \multicolumn{3}{c}{Adversarial}\\
    \cline{3-11}
    & & Accuracy & F1-Score & Yes & Accuracy & F1-Score & Yes & Accuracy & F1-Score & Yes \\
    \midrule
    LLaVA        & 7B & 72.16  & 78.22 & 76.29 & 61.37 & 71.52 & 85.63 & 58.67 & 70.12 & 88.33\\
    mPLUG-Owl    & 7B & 53.97 & 68.39 & 95.63 & 50.90 & 66.94 & 98.57 & 50.67 & 66.82 & 98.67\\
    MiniGPT-4    & 13B& 79.67 & 80.17 & 52.53 & 69.73 & 73.02 & 62.20 & 65.17 & 70.42 & 67.77\\
    InstructBLIP & 13B& 88.57 & \underline{89.27} & 56.57 & 82.77 & 84.66 & 62.37 & 72.10 & 77.32 & 73.03\\ 
    Shikra       & 7B & 86.90 & 86.19 & 43.26 & 83.97 & 83.16 & 45.23 & 83.10 & 82.49 & 46.50\\
    GroundingGPT & 7B & \underline{89.79}  & 89.22 & 43.13 & 88.23 & \underline{87.38} & 43.23 & 86.17 & 85.50 & 45.43\\ \midrule
    \multirow{2}{*}{TinyGroundingGPT} & 3B & \textbf{89.93}  & \textbf{89.47} & 43.08 & \underline{88.56} & \textbf{87.90} & 43.43 & \textbf{86.77} & \textbf{86.22} & 45.26\\
    ~ & 1.5B & 89.59  & 88.98 & 42.92 & \textbf{88.67} & \textbf{87.90} & 42.87 &\underline{86.74}  & \underline{86.04} & 44.77\\
    \bottomrule
    \end{tabular}}
    \caption{Results on the POPE benchmark for object hallucination evaluation. "Yes" represents the probability of positive answers to the given question.}
    \label{tab:pope}
\end{table*}

\subsection{Image Understanding Evaluation}
We evaluated TinyGroundingGPT on seven benchmarks, providing a comprehensive assessment of its performance across various metrics. Experimental results in Table~\ref{tab:image_understanding} show that TinyGroundingGPT-3B achieves results comparable to models such as MiniCPM-V-2, InternVL-2, which use dynamic high resolution or enriched training data. Compared to models with similar fine-tuning data, including LLaVA-1.5, GroundingGPT, TinyLLaVA and LLaVA-Phi, TinyGroundingGPT-3B demonstrates superior image understanding capabilities on the $\text{VQA}^{\text{v2}}$, GQA, SQA and POPE benchmarks, achieving increases of $2.6\%$ on MMB and $1.2\%$ on GQA over GroundingGPT-7B. Notably, TinyGroundingGPT-1.5B outperforms LLaVA-Phi, despite its larger parameter count, on most benchmarks. Overall, TinyGroundingGPT, optimized by our multi-scale visual knowledge alignment method, achieved impressive results across multiple evaluation sets.

\subsection{Object Hallucination Evaluation}
We evaluated MLLMs for object hallucination, as shown in Table~\ref{tab:pope}. Higher accuracy and F1-score metrics, along with a lower 'Yes' metric, indicate better performance. Our TinyGroundingGPT yielded outstanding results across all three sampling subsets. Notably, TinyGroundingGPT-3B outperformed larger models like InstructBLIP-13B in the challenging Adversarial subset, achieving an increase of $14.67\%$ in accuracy and a $8.90\%$ increase in F1 score, despite a decrease of $27.77\%$ in the 'Yes' metric. Compared to GroundingGPT-7B, our 3B model excelled in the Popular and Adversarial subsets for both accuracy and F1 score. Similarly, TinyGroundingGPT-1.5B achieved higher accuracy and F1 score than some larger models like Shikra while maintaining a lower 'Yes' score. This superior performance can be attributed to its fine-grained knowledge alignment from both global and local perspectives during training.

\subsection{Ablation Study}
\noindent{\textbf{Ablation Study on Additional Multi-scale Fine-grained Grounding Dataset.}} In Stage 2, compared to the traditional methods that rely solely on alignment datasets for object texts and coordinates, we utilized our constructed additional multi-scale fine-grained grounding datasets for TinyGroundingGPT, which enables us to achieve multi-scale alignment among object texts, images, and coordinates. The ablation study presented in Table~\ref{ablation1} shows that our proposed multi-scale fine-grained alignment outperformed the traditional referring data that only aligns object texts with coordinates. Whether for the 3B or 1.5B TinyGroundingGPT, our method enhanced performance on the RefCOCO, RefCOCO+, and RefCOCOg benchmarks. For instance, on the RefCOCO+ benchmark, there was an increase of $1.67\%$ for the 3B model and an increase of $0.87\%$ for the 1.5B model, demonstrating the effectiveness of our proposed fine-grained alignments and datasets.\\

\begin{table}[!ht]\small
    \centering
    \renewcommand\arraystretch{1.2}
    \setlength{\tabcolsep}{1mm}  {
    \begin{tabular}{ccccc}
    \hline
        Size & Multi-scale Align&  RefCOCO  &  RefCOCO+ &  RefCOCOg \\ \hline
        \multirow{2}{*}{3B} & \usym{2613} & 87.35 & 78.89 & 83.25 \\ 
        ~ & \checkmark & 88.50 & 80.56 & 83.69 \\ \hline
        \multirow{2}{*}{1.5B} & \usym{2613} & 85.93 & 77.05 & 79.54 \\ 
        ~ & \checkmark & 86.33 & 77.92 & 79.96 \\ \hline
    \end{tabular}}
    \caption{Ablation study on our Additional Multi-scale Fine-grained Grounding Dataset in Stage 2. If the model is trained without multi-scale alignment, it indicates that we are using only traditional referring data (which aligns only texts and coordinates). We report the average accuracy for each benchmark.}
    \label{ablation1}
\end{table}

\noindent{\textbf{Ablation Study on Global Grounding Datasets.}} In Stage 3, we utilized our constructed Global Grounding Datasets for TinyGroundingGPT to bridge different objects with varied representations and enhance global image comprehension. To evaluate the effectiveness of this strategy, we conducted an ablation study, with the results presented in Table~\ref{ablation2}. Notably, the results on the POPE benchmarks demonstrated a reduction in hallucinations. Overall, significant improvements in visual understanding benchmarks underscored the value of detailed global knowledge learning and the Global Grounding Datasets, which enhanced global object alignment by connecting different objects represented by texts, coordinates, and images through multi-round grounding conversations and grounding descriptions.
\begin{table}[!ht]\small
    \centering
    \renewcommand\arraystretch{1.2}
    \setlength{\tabcolsep}{1mm}  {
    \begin{tabular}{ccccccc}
    \hline
        Size & Global Align &  GQA & $\text{VQA}^{\text{v2}}$ & SQA & POPE & MMB \\ \hline
        \multirow{2}{*}{3B} & \usym{2613} & 61.7 & 77.4  & 65.6 & 86.6 & 63.1\\ 
        ~ & \checkmark & 63.3 & 79.3 & 70.3 & 87.9 & 66.4\\ \hline
        \multirow{2}{*}{1.5B} & \usym{2613} & 60.3 & 77.3  & 62.1 & 86.4 & 63.0 \\ 
        ~ & \checkmark & 62.2 & 77.9 & 63.1 & 87.6 & 64.2 \\ \hline
    \end{tabular}}
    \caption{Ablation study on our Global Grounding Datasets in Stage 3. If the model is trained without global alignment, it indicates that we do not use these datasets to further align different objects represented by texts, coordinates, and images.}
    \label{ablation2}
\end{table}

\noindent{\textbf{Ablation Study on Models with Larger Parameter.}} We applied our method to TinyGroundingGPT with the larger language model Qwen2.5-7B, as illustrated in Appendix~\ref{7b_ref}. The results show an average increase of $0.47\%$ across all grounding task benchmarks, highlighting the effectiveness and generalizability of our proposed method.

\begin{figure}[!htp]
\centering
\includegraphics[width=1\columnwidth]{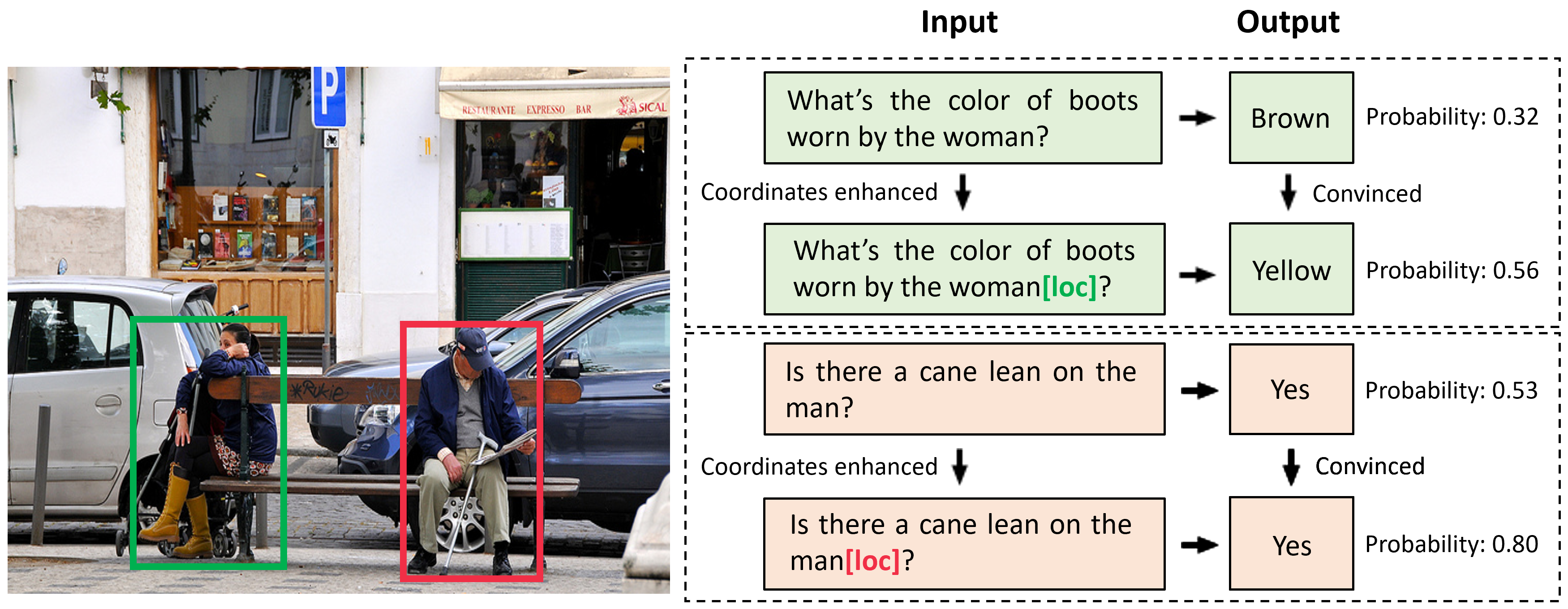}\\
\caption{A case for the outputs of our TinyGroundingGPT when the input is either enhanced with coordinates or not. Probability values indicate the likelihood of generating corresponding tokens.}
\label{convince_case}
\end{figure}

\begin{figure*}[!htp]
\centering
\includegraphics[width=2\columnwidth]{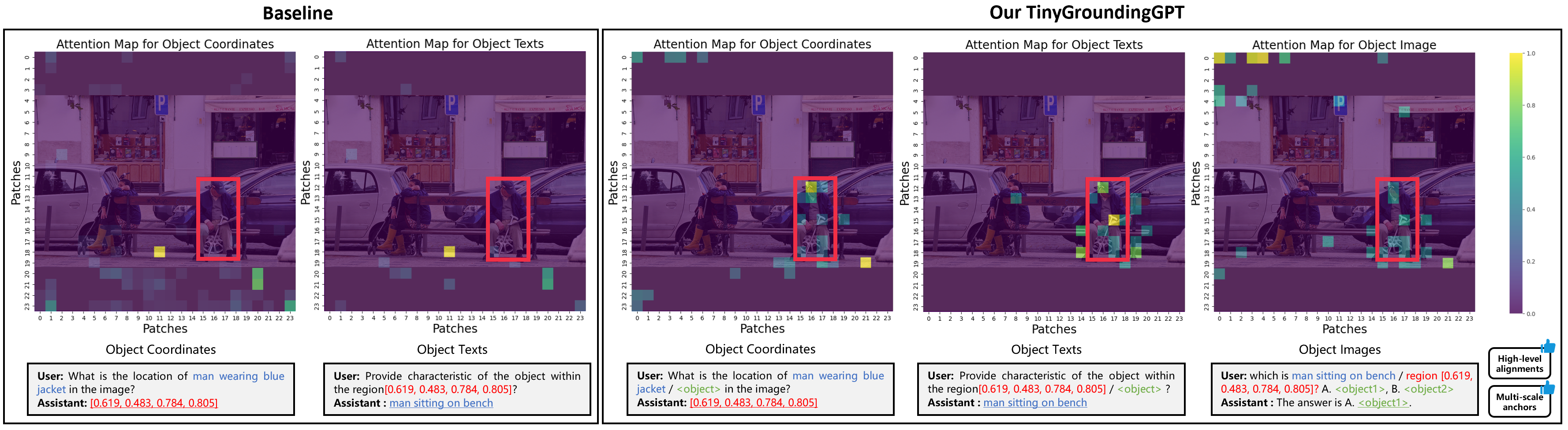}\\
\caption{Visualization of the attention map for image patches with different object representation outputs (texts, coordinates, and images, underlined). The red bounding box denotes the target region. The attention at the four corners serves as anchors for grounding, while attention at specific objects highlights their importance.}
\label{visualization}
\end{figure*}

\section{Discussion}
\label{sec:discussion}
\subsection{Effectiveness of fine-grained knowledge}\label{assess}
In addition to improved grounding ability, our proposed fine-grained visual knowledge alignment method also enhances comprehensive image understanding. We provide examples of image descriptions generated by TinyGroundingGPT in Appendix Fig.~\ref{cases}. Notably, our model, optimized through multi-granularity knowledge alignment, effectively avoids incorrect or nonexistent object descriptions. To further assess this annotation ability, we selected 50 images from RefCOCO-test and utilized GPT-4V to score image descriptions produced by different methods. As detailed in Appendix~\ref{assess_appendix}, TinyGroundingGPT achieves better overall quality and richness compared to GroundingGPT-7B and Qwen2-VL-2B. Additionally, this aligned fine-grained knowledge can be used to enhance input questions for TinyGroundingGPT, resulting in more convincing and certain responses. As shown in Fig.~\ref{convince_case}, compared to directly asking questions about objects in an image, enhancing the object texts with coordinates in the queries yields more accurate and persuasive responses from our TinyGroundingGPT. This finding provides insights for further unleashing the potential of fine-grained MLLMs.

\subsection{Interpretability for high-level alignments}\label{Interpretability}
Grounding MLLMs fundamentally model the maximum likelihood output based on visual inputs and text prompts. By conditioning on the referring prompt, the model identifies which parts of the image significantly influence the output. Consequently, the attention map in grounding MLLMs not only enhances interpretability but also illustrates the alignment between the model's output and the input image. To demonstrate the effectiveness of our multi-scale fine-grained grounding capability, we visualize the attention map in the last layer of our TinyGroundingGPT. The process for obtaining the heatmap of attention involves several steps: (1) we select the attention scores between image patches and object representations (i.e., texts, coordinates, and images); (2) we sum the attention scores across the dimensions of both the attention heads and object representations; (3) We map the normalized attention scores onto the input image patches. 

As shown in Fig.~\ref{visualization}, the attention maps of our TinyGroundingGPT reveal distinct location attributions, unlike the baseline GroundingGPT-7B. For object coordinates, high attention scores are concentrated at the four corners of the image, serving as anchors for bounding box coordinates, as well as at the locations of the intended objects mentioned in the prompt. When prompted to describe a specific region, the model directs increased attention to the corresponding object patches. For the output of an object image, the attention values between image patches and the target object highlight relevant regions and reinforce grounding anchors. This indicates that TinyGroundingGPT effectively learns both aligned features and grounding information for object images. In summary, our findings underscore the effectiveness of the proposed fine-grained visual knowledge alignment method, achieving high-level alignment among different object representations. This provides insights for further explaining MLLMs, particularly in grounding tasks. More visualizations can be found in Appendix Fig.~\ref{more_visualization}.

\section{Conclusion}
\label{sec:conclusion}
In this paper, we introduce a novel fine-grained visual knowledge alignment method for MLLMs to address the limitations of fine-grained alignments in previous works. Our method progresses from easy to hard, emphasizing multi-scale fine-grained alignments among object texts, coordinates, and images from both local and global perspectives. This empowers models to effectively learn fine-grained knowledge and facilitates reasoning and grounding tasks. Additionally, we develop a multi-scale fine-grained enhancement data synthesis pipeline that leverages open-source datasets and advanced models to generate over 300K essential training samples. Building on this foundation, we train TinyGroundingGPT, a series of smaller models (1.5B and 3B parameters) optimized through high-level alignments, capable of handling various visual and grounding tasks, often surpassing larger models. Experimental results demonstrate the effectiveness of our proposed method and the generated datasets. Our work contributes to the advancement of practical applications for MLLMs.

{\small
\bibliographystyle{ieeenat_fullname}
\bibliography{main}
}
\newpage
\section{Appendix}
\label{sec:Appendix}

\subsection{Implementation details}\label{Implementation}
We present additional details about our experimental configuration to facilitate the reproduction of our model. The hyperparameters for all stages are summarized in Table~\ref{training_details}.

\begin{table}[H]\small
    \centering
    \renewcommand\arraystretch{1.2}
    \setlength{\tabcolsep}{1mm}
    \begin{tabular}{c|cccc}
    \hline
        \multirow{2}{*}{Size} & \multicolumn{2}{c}{Stage 1} & \multirow{2}{*}{Stage 2} & \multirow{2}{*}{Stage 3} \\
        & Pretrain & Finetune & & \\ \hline
        Batch size & 32 & 32 & 32 & 16 \\ 
        Learning rate & 1e-3 & 2e-5 & 2e-5 & 2e-5 \\ 
        Epochs & 1 & 1 & 1 & 2 \\ 
        Learning schedule & \multicolumn{4}{c}{Cosine decay} \\ 
        Warm-up ratio & 0.03 & 0.03 & 0.03 & 0.03 \\ 
        Weight decay & 0 & 0 & 0 & 0 \\ 
        BF16 & \checkmark & \checkmark & \checkmark & \checkmark \\
        TF32 & \checkmark & \checkmark & \checkmark & \checkmark \\
        DeepSpeed stage & \multicolumn{4}{c}{ZeRO2} \\ 
        GPUs & \multicolumn{4}{c}{8xA100} \\  \hline
    \end{tabular}
    \caption{The hyperparameters for model training.}
    \label{training_details}
\end{table}

\subsection{Dataset details}\label{Dataset}
\begin{figure*}[htp]
\centering
\includegraphics[width=1.8\columnwidth]{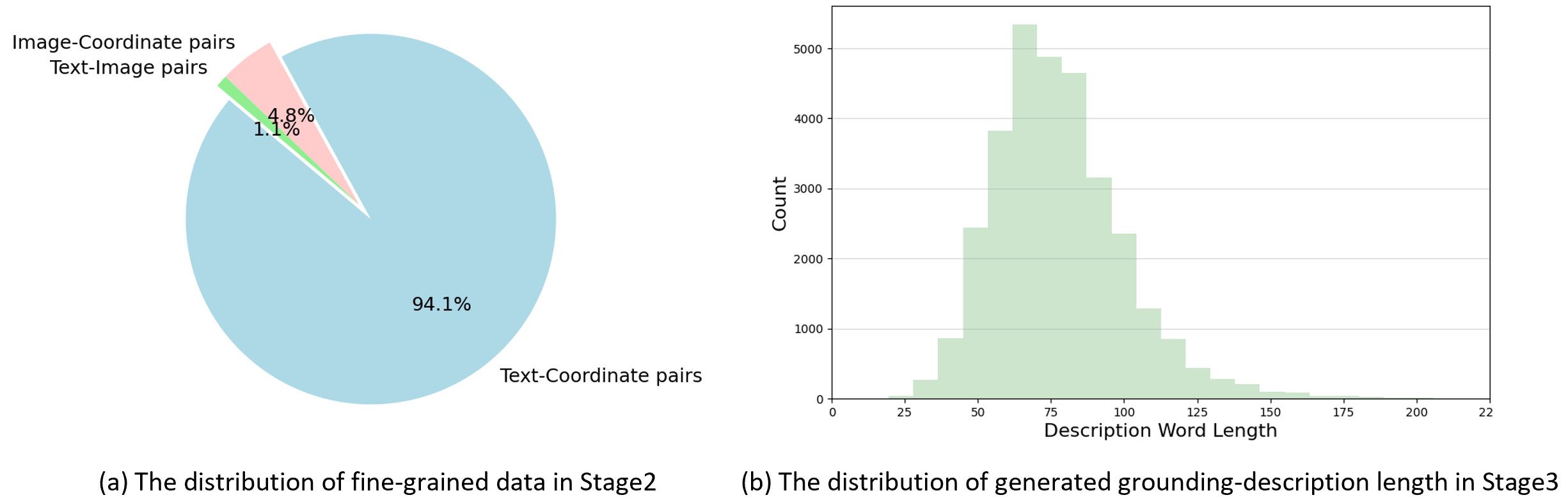}\\
\caption{The distribution of fine-grained data in Stage2 and generated grounding-description length in Stage3.}
\label{distribution}
\end{figure*}

We provide additional details about the datasets we utilized, as summarized in Table~\ref{training_data}.  Moreover, the distribution of fine-grained data and image description length are shown in Fig.~\ref{distribution}. We also include additional examples of the generated datasets in Fig.~\ref{QApairs} for Stage 2 and in Fig.~\ref{QApairs2} for Stage 3. As described in Section~\ref{datapipeline}, we developed a multi-scale fine-grained enhancement data synthesis pipeline, which includes the construction of a multi-scale fine-grained grounding dataset (in Stage 2) and a global grounding dataset (in Stage 3). In Fig.~\ref{prompt1}, we present the prompt messages used for object recognition and relation extraction to prepare additional data material. Fig.~\ref{prompt2} illustrates the detailed processing steps involved in constructing the multi-scale fine-grained grounding dataset. Furthermore, Figs.~\ref{prompt3} and \ref{prompt4} outline the processing steps for constructing the global grounding dataset. 

\begin{table}[!htp]\small
    \centering
    \renewcommand\arraystretch{1.2}
    \setlength{\tabcolsep}{1mm}
    \begin{tabular}{c|c|c|c}
    \toprule
       Stage & \multicolumn{2}{c}{Dataset} & Samples \\ \midrule
       Stage1 & \multicolumn{2}{c}{LLaVA-Pretrain-595k} & 595K \\
       \cline{2-4}
       \multirow{3}{*}{Stage2} & \multirow{3}{*}{Fine-grained data}  & Text-coordinate pairs & 4.1M\\
        &  & Image-coordinate pairs & 210K\\
         &  & Text-image pairs & 46K\\
        \cline{2-4}
        \multirow{3}{*}{Stage3} & \multirow{3}{*}{SFT data}  &  LLaVA-v1.5, ShareGPT4V & 665K\\
        &  & Grounding-conv & 40K\\
         &  & Grounding-description & 32K\\
         \bottomrule
    \end{tabular}
    \caption{The dataset details used for model training.}
    \label{training_data}
\end{table}

\subsection{Grounding ability for larger model}\label{7b_ref}
We further apply our proposed fine-grained visual knowledge alignment method to TinyGroundingGPT, utilizing Qwen2.5-7B as the language model with larger parameters, to evaluate its image grounding capability. The results, summarized in Table~\ref{tab:refcoco-7b}, demonstrate an average increase of $0.47\%$ across all grounding task benchmarks. Specifically, on RefCOCO-testB, accuracy improves by $1.01\%$, highlighting the effectiveness of our proposed method.

\begin{table*}[ht]\small
    \centering
    \begin{tabular}{cc|ccc|ccc|cc|c}
    \toprule  
    \multirow{2}{*}{Model} & \multirow{2}{*}{Multi-scale Align} & \multicolumn{3}{c}{RefCOCO} & \multicolumn{3}{c}{RefCOCO+} & \multicolumn{2}{c}{RefCOCOg} &
     \multirow{2}{*}{Avg}\\
    \cline{3-10}  & & val & testA & testB & val & testA & testB & val & test & \\
    \midrule
    \multirow{2}{*}{TinyGroundingGPT-7B} & \usym{2717} & 90.28 & 92.62 & 86.47 & 83.98 & 88.08 & 77.90 & 85.27 & 85.43 & 86.25\\

    & \usym{2713} & 90.72 & 92.31 & 87.48 & 84.56 & 88.76 & 78.71 & 85.36 & 85.86 & \textbf{86.72(+0.47)}\\
    \bottomrule
    \end{tabular}
    \caption{Performance comparison on the referring expression comprehension(REC) task for whether conducting our proposed Multi-scale Fine-grained Local Knowledge Alignment.}
    \label{tab:refcoco-7b}
\end{table*}

\subsection{Assessment for image annotation}\label{assess_appendix}
As illustrated in Section~\ref{assess}, we provided examples of image descriptions generated by TinyGroundingGPT in Fig.~\ref{cases}. Moreover, we selected 50 images from RefCOCO-test and utilized GPT-4V to evaluate image descriptions produced by various methods. We assessed the image descriptions using scores ranging from 1 to 5 across three perspectives: "Quality," which reflects overall quality; "Richness," which measures the diversity of object descriptions; and "Accuracy," which pertains to precision. The prompt used for GPT-4V and the scoring details are presented in Table~\ref{prompt_GPT4V}. As the results summarized in Table~\ref{scores}, TinyGroundingGPT achieved better overall quality and richness compared to GroundingGPT-7B and Qwen2-VL-2B.

\begin{table*}[!htp]\small
    \renewcommand\arraystretch{1.5}
    \centering
    
    \begin{tabular}{p{5cm} p{5cm} p{5cm}}
    \hline
    \multicolumn{3}{c}{\textbf{Evaluate the image description based on the following criteria:}} \\ \midrule
    \textbf{Quality (1-5):} & \textbf{Richness (1-5):} & \textbf{Accuracy (1-5):} \\
    1 - The description is incoherent, lacks flow, and does not effectively convey the contents of the image. & 1 - The description only mentions a few basic objects or elements in the image, without any contextual details or relationships. & 1 - The description contains multiple significant inaccuracies or errors in identifying objects, elements, or their characteristics. \\
    2 - The description has some coherence but is still disjointed, with limited flow and incomplete coverage of the image. & 2 - The description includes some additional details about the objects or elements but lacks depth in terms of their relationships or broader context. & 2 - The description has some inaccuracies or errors in identifying objects, elements, or their characteristics. \\
    3 - The description is generally coherent, with reasonable flow, and covers most of the key elements in the image. & 3 - The description provides a reasonable level of detail about the objects and elements, as well as some of their relationships or broader context. & 3 - The description is generally accurate in identifying the objects, elements, and their characteristics, with only minor inaccuracies. \\
    4 - The description is coherent, with good flow, and comprehensively covers the important aspects of the image. & 4 - The description is rich in detail, covering a diverse range of objects, elements, their relationships, and the broader context of the scene. & 4 - The description is highly accurate in identifying the objects, elements, and their characteristics, with minimal to no inaccuracies. \\
    5 - The description is highly coherent, with excellent flow, and articulately captures the essence of the image in a compelling manner. & 5 - The description is exceptionally rich, providing abundant details about the diverse array of objects, elements, their intricate relationships, and the comprehensive context of the scene. & 5 - The description is completely accurate in identifying all the objects, elements, and their characteristics, with no discernible errors or hallucinations. \\ \hline
    \end{tabular}
    \caption{The prompt for GPT-4V to assess descriptions from the perspectives of Quality, Richness, and Accuracy.}
    \label{prompt_GPT4V}
    
\end{table*}

\begin{table}[H]\small
    \centering
    \renewcommand\arraystretch{1.2}
    \setlength{\tabcolsep}{0.6mm}{
    \begin{tabular}{cccc}
    \hline
        Model  & Quality & Richness  & Accuracy\\ \midrule
        GPT-4V & 4.24 & 4.10 & 4.88\\
        GroundingGPT-7B & 3.68 & 3.20 & 3.38\\
        Qwen2-VL-2B  & 3.90 & 3.64 & 4.18\\ \midrule
        TinyGroundingGPT-3B & 4.04 & 3.90 & 3.66\\ \hline
    \end{tabular}}
    \caption{The assessment for image annotation by GPT-4V includes "Quality" for overall quality, "Richness" for the diversity of object descriptions, and "Accuracy" for precision. Scores are based on the average ratings (1-5) from 50 samples. "Coordinates" denotes whether outputting object texts with coordinates.}
    \label{scores}
\end{table}

\subsection{More visualizations}
As illustrated in Section~\ref{Interpretability}, we visualized the last-layer attention maps of both the GroundingGPT-7B baseline and our TinyGroundingGPT-1.5B. Additional visualizations are displayed in Fig.~\ref{more_visualization}. As shown, TinyGroundingGPT reveals more distinct location attributions, indicating that it effectively learned multi-scale fine-grained knowledge and achieved high-level alignments among object texts, coordinates, and images. This provides insights for further explaining MLLMs, particularly in grounding tasks. We also provide a demo for utilizing TinyGroundingGPT in Fig.~\ref{demo}.

\begin{figure*}[htp]
\centering
\includegraphics[width=1.4\columnwidth]{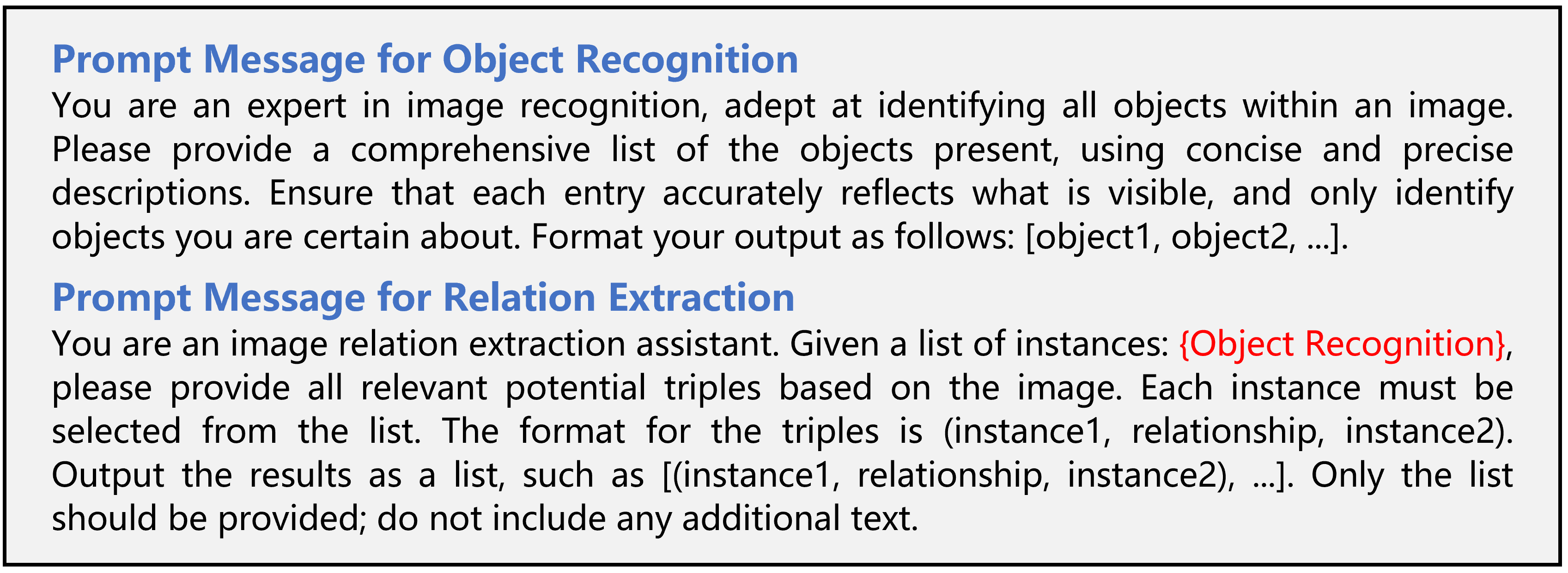}\\
\caption{The prompt message for object recognition and relation extraction.}
\label{prompt1}
\end{figure*}

\begin{figure*}[htp]
\centering
\includegraphics[width=2\columnwidth]{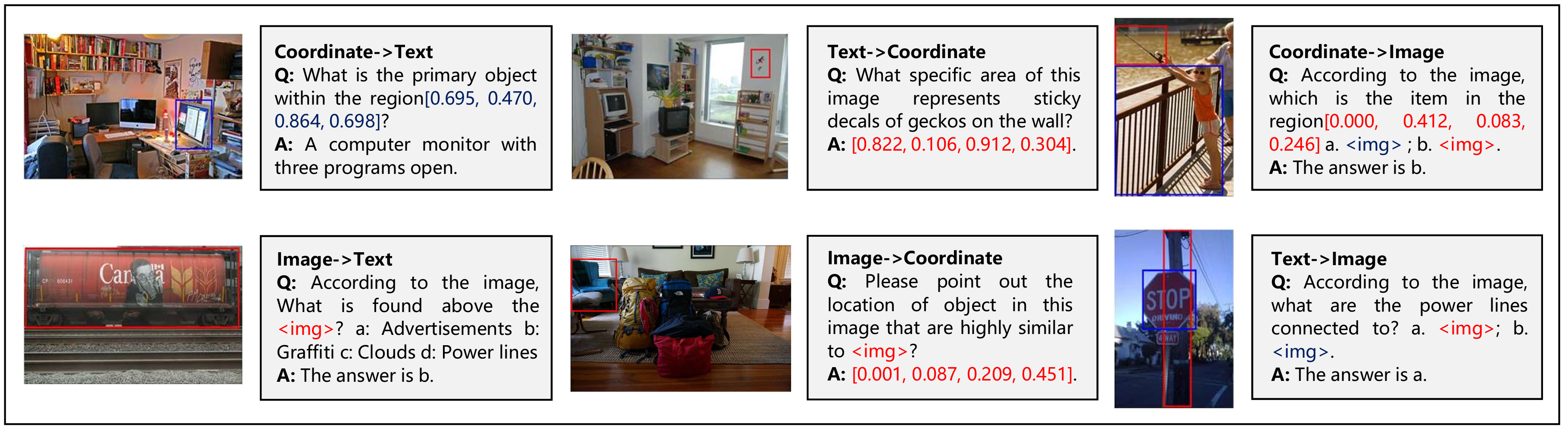}
\caption{Our generated various kinds of data used in Stage2 for achieving high-level alignments among texts, coordinates, and images, where the $<\text{img}>$ denotes the corresponding augmented object image.}
\label{QApairs}
\end{figure*}

\begin{figure*}[htp]
\centering
\includegraphics[width=2\columnwidth]{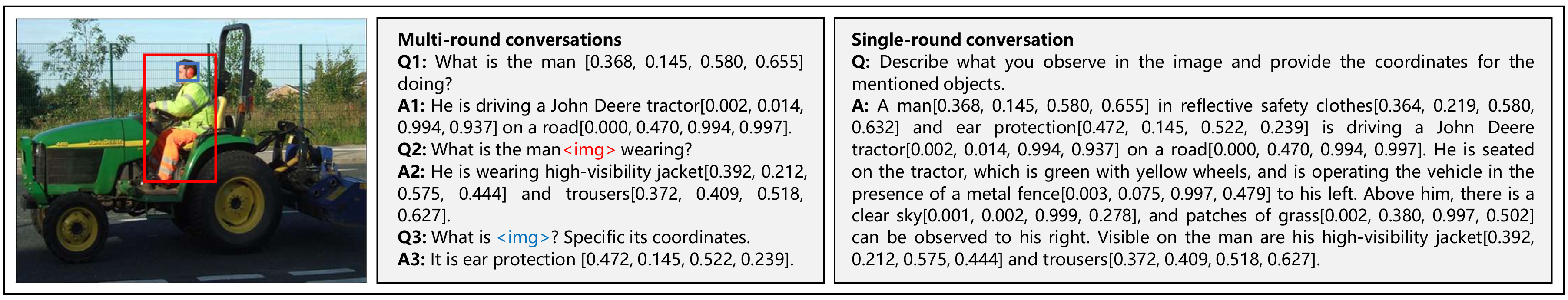}
\caption{Our generated various kinds of data used in Stage3 for achieving global object alignment, where the $<\text{img}>$ denotes the corresponding augmented object image.}
\label{QApairs2}
\end{figure*}

\begin{figure*}[htp]
\centering
\includegraphics[width=1.4\columnwidth]{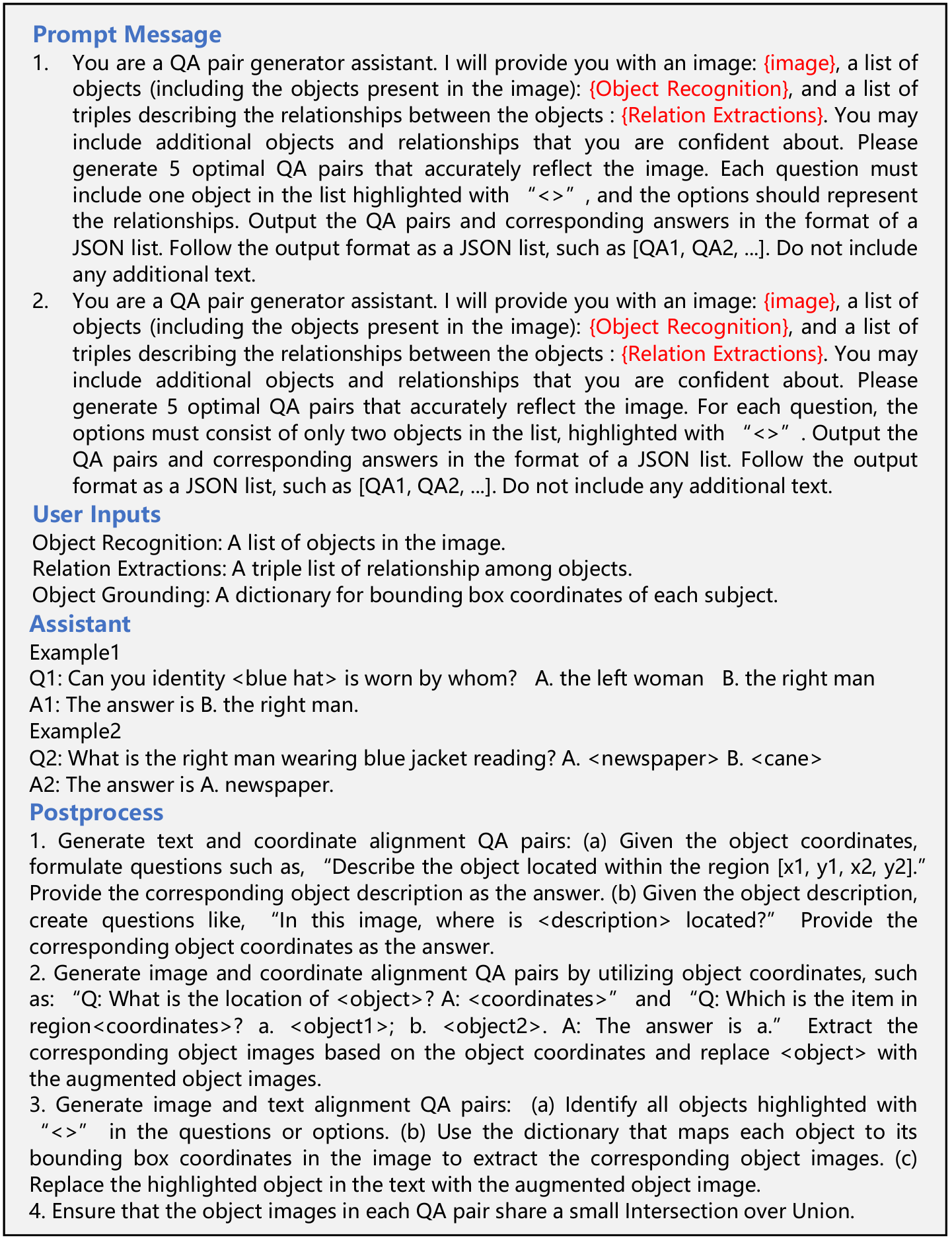}
\caption{The prompt message and user's input example used for generating our Fine-grained Grounding Dataset in Stage2.}
\label{prompt2}
\end{figure*}

\begin{figure*}[htp]
\centering
\includegraphics[width=1.4\columnwidth]{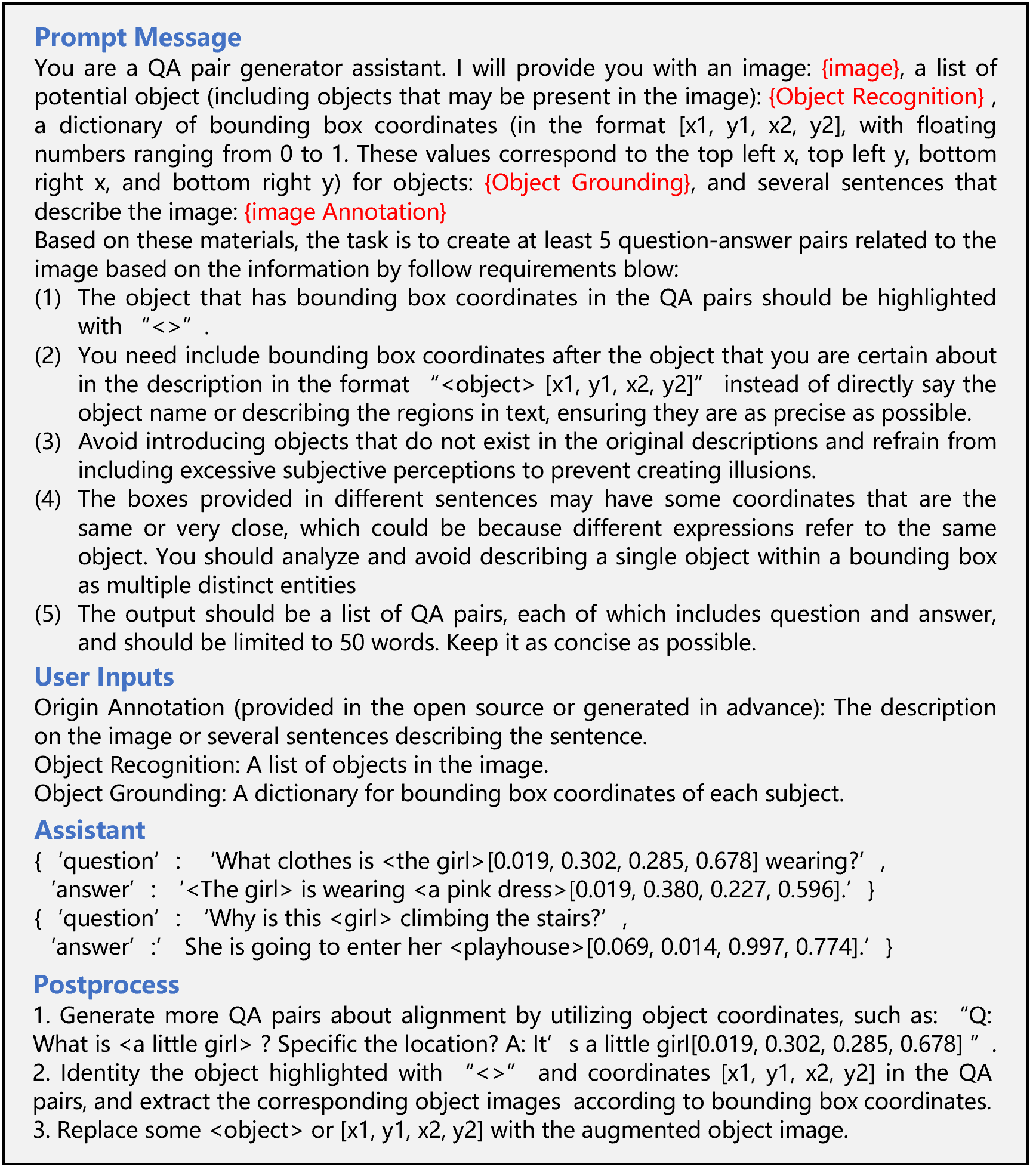}
\caption{The prompt message and user's input example used for generating our Multi-round Grounding Conversation Data in Stage3.}
\label{prompt3}
\end{figure*}

\begin{figure*}[htp]
\centering
\includegraphics[width=1.4\columnwidth]{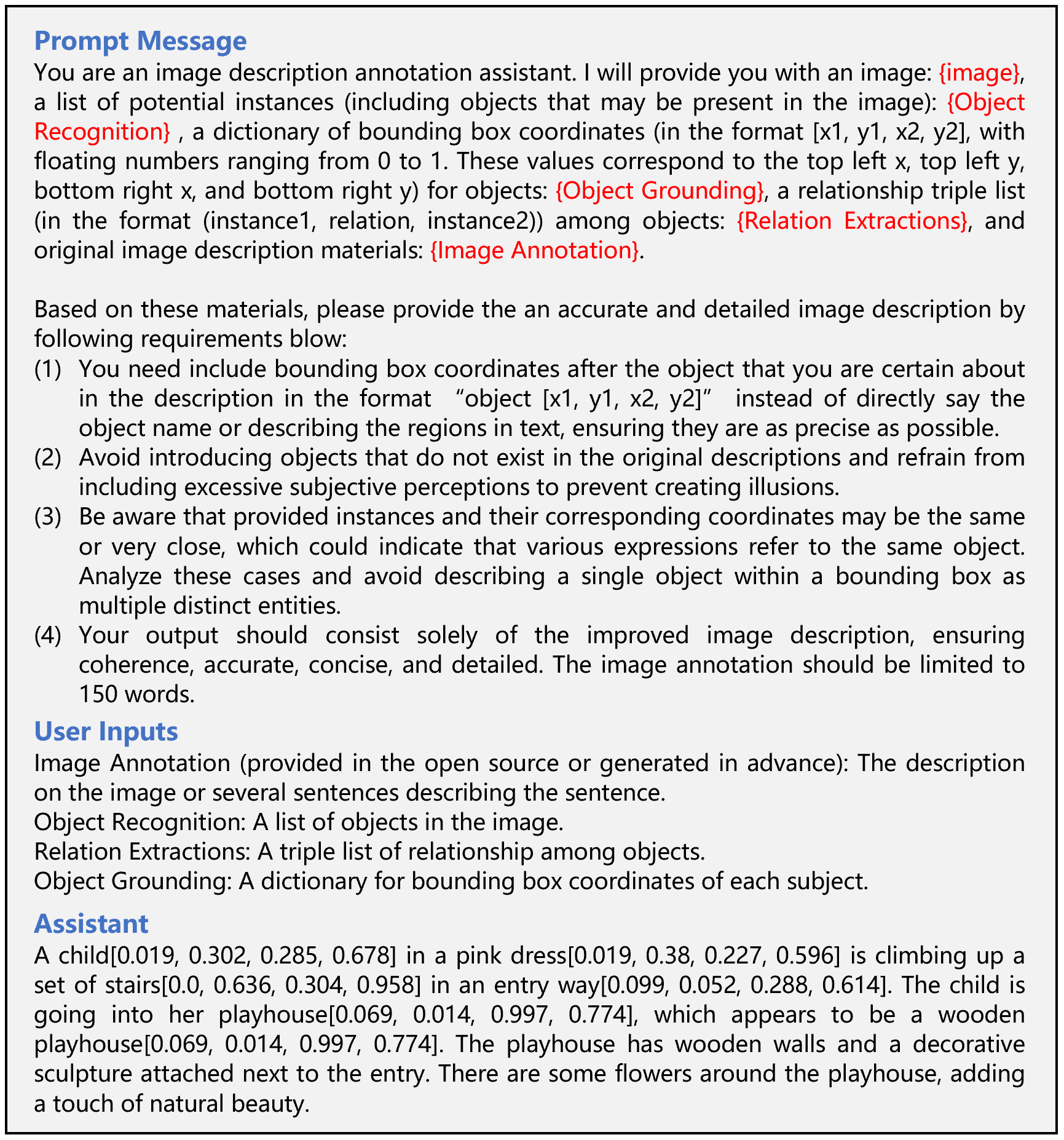}
\caption{The prompt message and user's input example used for generating our Grounding Description Data in Stage3.}
\label{prompt4}
\end{figure*}

\begin{figure*}[htp]
\centering
\includegraphics[width=2\columnwidth]{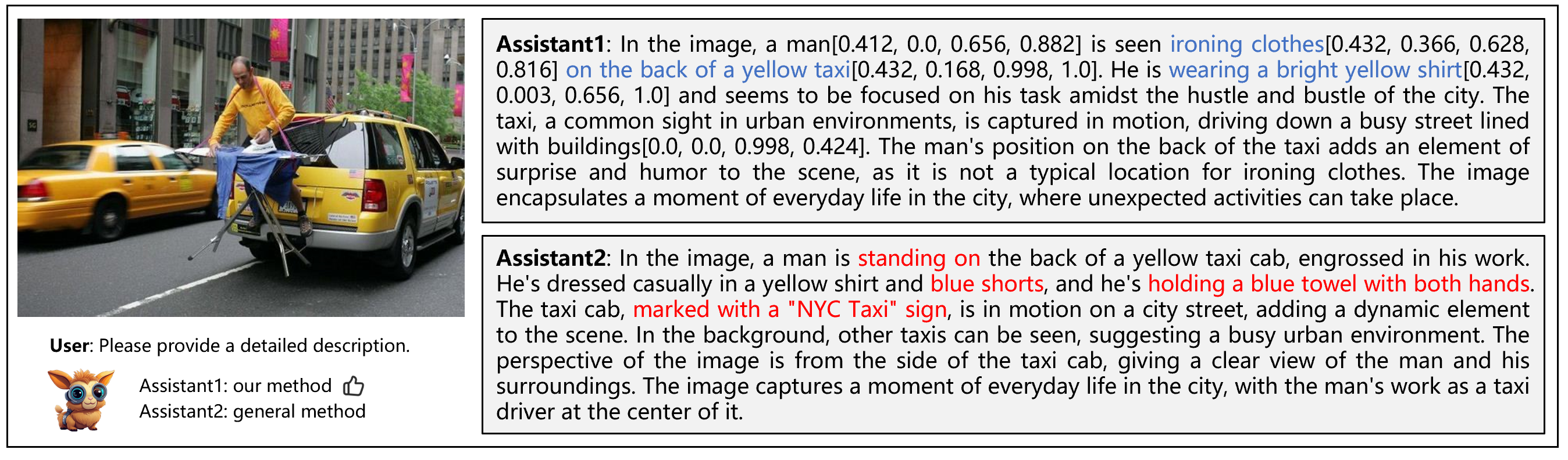}
\caption{A comparison of generated image descriptions between TinyGroundingGPT trained with our method and without it.}
\label{cases}
\end{figure*}

\begin{figure*}[htp]
\centering
\includegraphics[width=2\columnwidth]{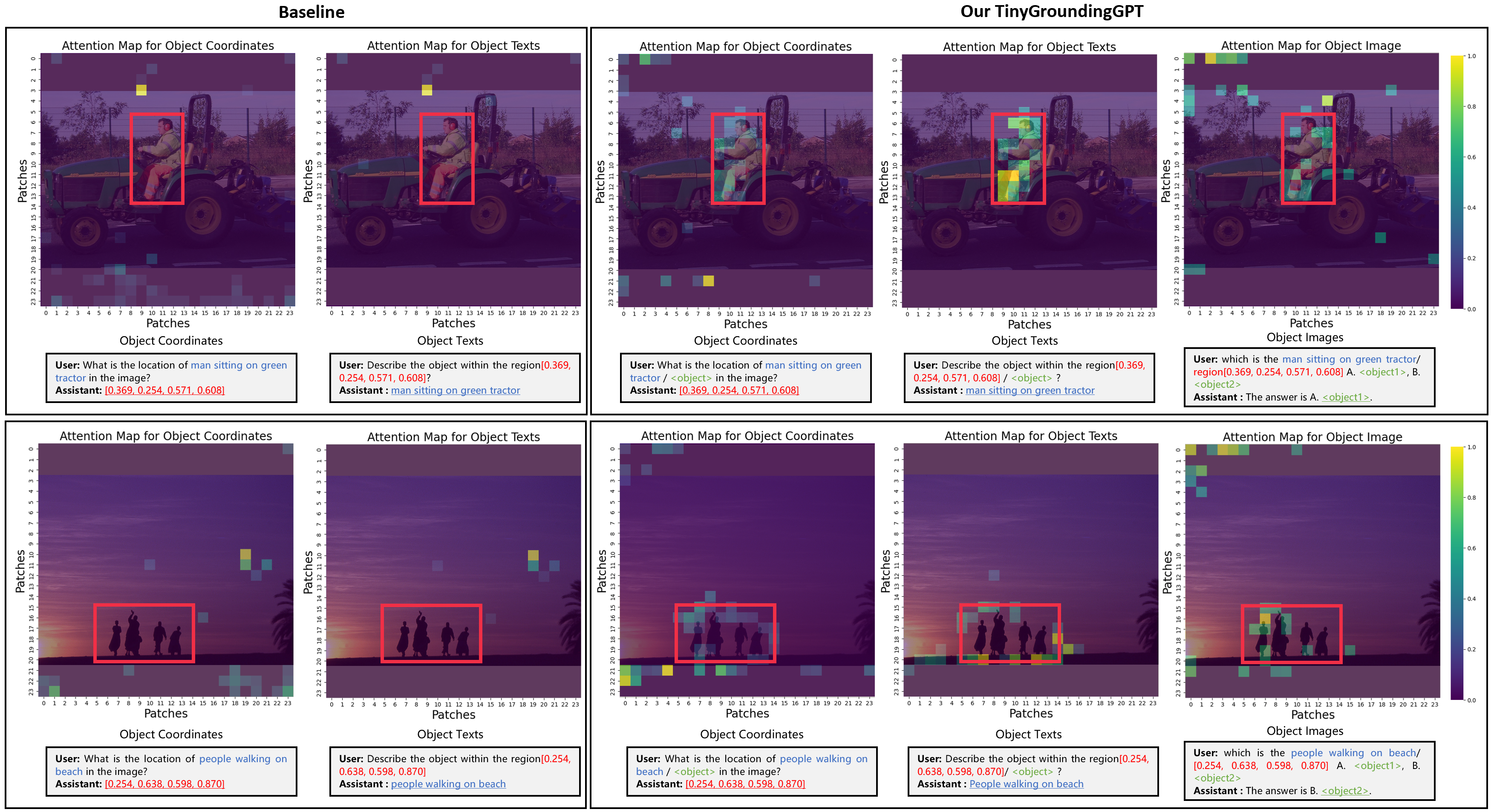}
\caption{The visualization of the attention map for image patches with different object representation outputs (texts, coordinates, and images, which are underlined), where the red bounding box denotes the target region.}
\label{more_visualization}
\end{figure*}

\begin{figure*}[htp]
\centering
\includegraphics[width=2\columnwidth]{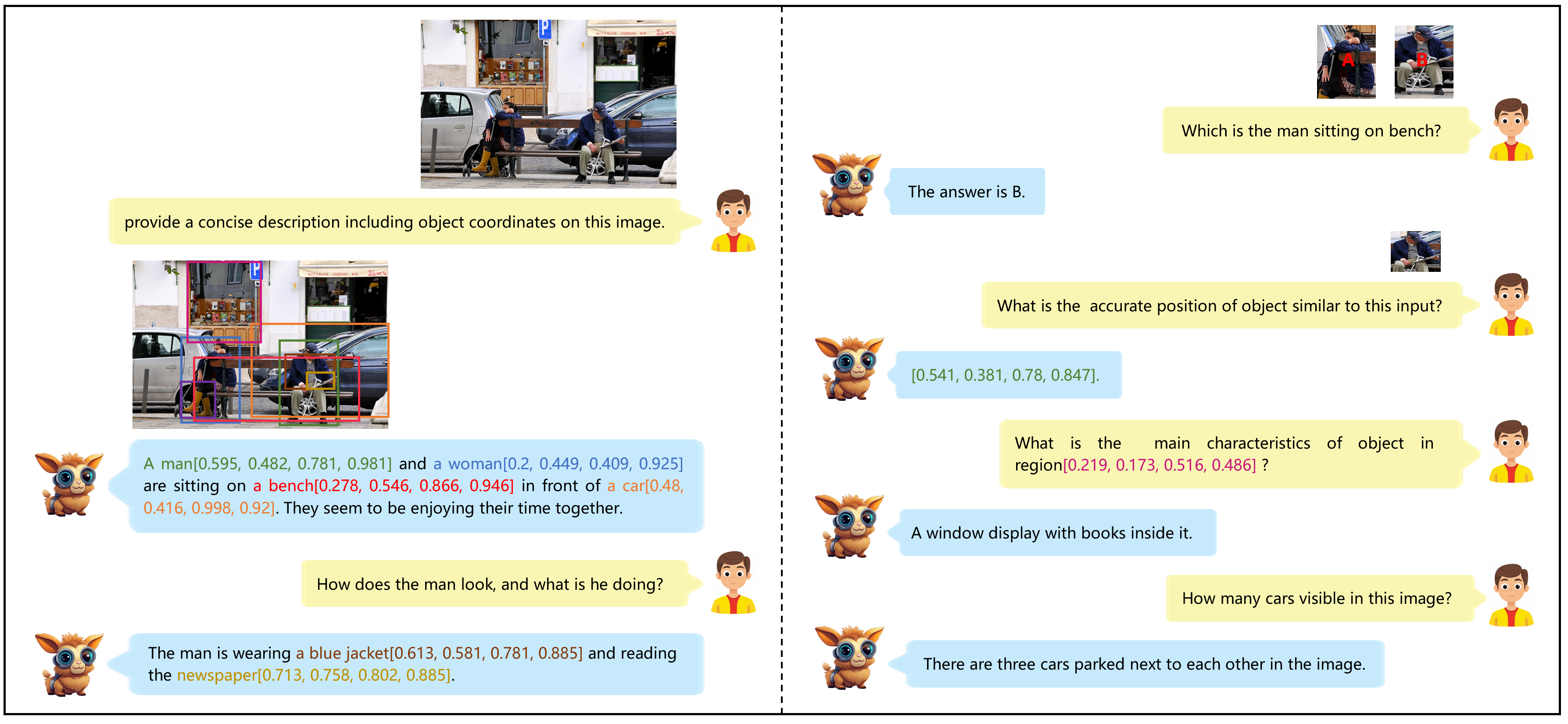}\\
\caption{A demo for the use of our TinygroundingGPT.}
\label{demo}
\end{figure*}

\end{document}